\documentclass{bmvc2k}
\usepackage{amsfonts}
\usepackage{amsmath}
\usepackage{amssymb} 
\usepackage{booktabs}
\usepackage{graphicx}
\usepackage{microtype}
\usepackage{nicefrac}
\usepackage{pifont}
\usepackage{soul}
\usepackage[capitalise]{cleveref}
\usepackage[bottom]{footmisc}
\usepackage{wrapfig}

\newif\ifdark
\IfFileExists{/Users/vedaldi/.bash_profile}{
\immediate\write18{%
if defaults read -g AppleInterfaceStyle 2>/dev/null;
then echo \\darktrue > /tmp/displaymode.tex;
else echo \\darkfalse > /tmp/displaymode.tex; fi}
\input{/tmp/displaymode.tex}
\ifdark
\definecolor{pcolor}{HTML}{1E1E1E}
\definecolor{tcolor}{HTML}{C5C5C5}
\else
\definecolor{pcolor}{HTML}{FDF6E3}
\definecolor{tcolor}{HTML}{333333}
\fi
\pagecolor{pcolor}
\color{tcolor}
\hbadness=\maxdimen
\vbadness=\maxdimen
\vfuzz=30pt
\hfuzz=30pt
}{}

\title{The Curious Layperson: Fine-Grained Image Recognition without Expert Labels}

\addauthor{Subhabrata Choudhury}{subha@robots.ox.ac.uk}{1}
\addauthor{Iro Laina}{iro@robots.ox.ac.uk}{1}
\addauthor{Christian Rupprecht}{chrisr@robots.ox.ac.uk}{1}
\addauthor{Andrea Vedaldi}{vedaldi@robots.ox.ac.uk}{1}

\addinstitution{
 Visual Geometry Group\\
 University of Oxford\\
 Oxford, UK
}
\runninghead{Choudhury, Laina, Rupprecht, Vedaldi}{The Curious Layperson}

\def\ie{\emph{i.e}\bmvaOneDot}
\def\eg{\emph{e.g}\bmvaOneDot}

\makeatletter
\renewcommand{\paragraph}{%
  \@startsection{paragraph}{4}%
  {\z@}{0.25em}{-1em}%
  {\normalfont\normalsize\bfseries}%
}
\makeatother

\makeatletter
\newcommand{\cmark}{\ding{51}}%
\newcommand{\xmark}{\ding{55}}%
\makeatother

\newcommand{\ourtask}{CLEVER\xspace}

\DeclareMathOperator{\bert}{T}

\begin{document}

\maketitle

\begin{abstract}
Most of us are not experts in specific fields, such as ornithology.
Nonetheless, we do have general image and language understanding capabilities that we use to match what we see to expert resources. 
This allows us to expand our knowledge and perform novel tasks without ad-hoc external supervision.
On the contrary, machines have a much harder time consulting expert-curated knowledge bases unless trained specifically with that knowledge in mind.
Thus, in this paper we consider a new problem: fine-grained image recognition without expert annotations, which we address by leveraging the vast knowledge available in web encyclopedias. 
First, we learn a model to describe the visual appearance of objects using non-expert image descriptions.
We then train a fine-grained textual similarity model that matches image descriptions with documents on a sentence-level basis.
We evaluate the method on two datasets %
and compare with several strong baselines and the state of the art in cross-modal retrieval\@.
Code is available at: \url{https://github.com/subhc/clever}.
\end{abstract}

\section{Introduction}\label{s:intro}

\begin{figure}[t]
\center
\includegraphics[width=0.95\linewidth, trim=0cm 5.9cm 0.0cm 0cm, clip]{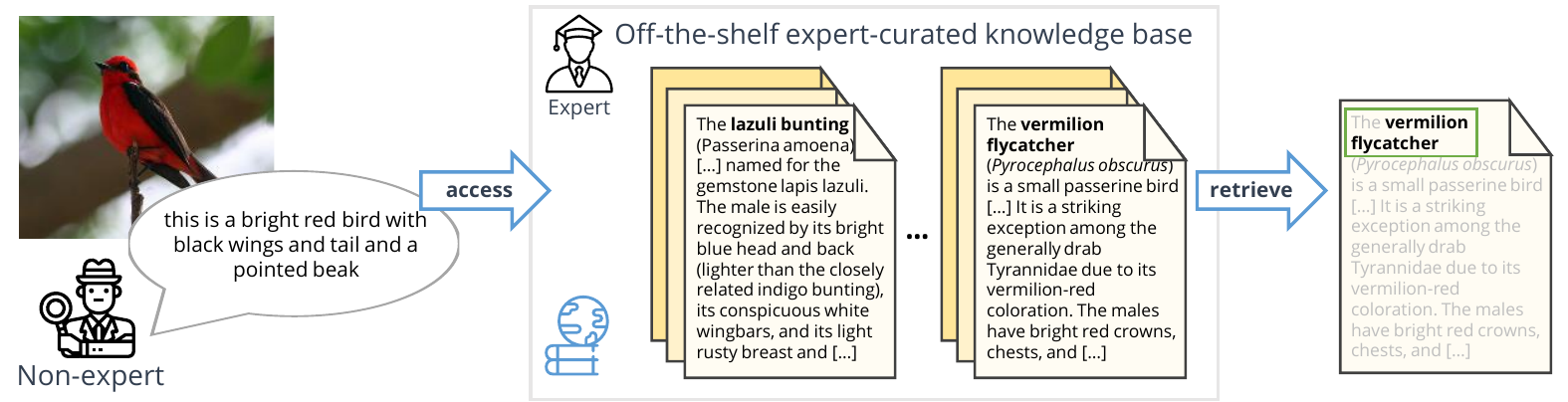}
\caption{\textbf{Fine-Grained Image Recognition without Expert Labels.} We propose a novel task that enables fine-grained classification without using expert class information (\eg bird species) during training. We frame the problem as document retrieval from general image descriptions by leveraging existing textual knowledge bases, such as Wikipedia.}\label{f:splash}
\label{fig:teaser}
\end{figure}

Deep learning and the availability of large-scale labelled datasets have led to remarkable advances in image recognition tasks, including fine-grained recognition~\cite{wah2011caltech, Nilsback06, van-horn17the-inaturalist}.
The problem of fine-grained image recognition amounts to identifying subordinate-level categories, such as different species of birds, dogs or plants. 
Thus, the supervised learning regime in this case requires annotations provided by domain \emph{experts} or citizen scientists~\cite{van2015building}.

While most people, unless professionally trained or enthusiasts, do not have knowledge in such specific domains, they are generally capable of consulting %
existing expert resources such as books or online encyclopedias, \eg Wikipedia.
As an example, let us consider bird identification. 
Amateur bird watchers typically rely on field guides to identify observed species. 
As a general instruction, one has to answer the question ``what is most noticeable about this bird?'' before skimming through the guide to find the best match to their observation. 
The answer to this question is typically a detailed description of the bird’s shape, size, plumage colors and patterns. 
Indeed, in \cref{fig:teaser}, the non-expert observer might not be able to directly identify a bird as a ``Vermillion Flycatcher'', but they \emph{can} simply describe the appearance of the bird: ``\textit{this is a bright red bird with black wings and tail and a pointed beak}''.
This description can be matched to an expert corpus
to obtain the species and other expert-level information.

On the other hand, machines have a much harder time consulting off-the-shelf expert-curated knowledge bases. %
In particular, most algorithmic solutions are designed to address a \emph{specific} task with datasets constructed \emph{ad-hoc} to serve precisely this purpose.
Our goal, instead, is to investigate whether it is possible to re-purpose general image and text understanding capabilities to allow machines to consult
already existing \emph{textual} knowledge bases to address a new task, such as recognizing a bird.  %

We introduce a novel task inspired by the way a layperson would tackle fine-grained recognition from visual input; we name this \textbf{\ourtask}, \ie \textbf{C}urious \textbf{L}ayperson-to-\textbf{E}xpert \textbf{V}isual \textbf{E}ntity \textbf{R}ecognition.
Given an image of a subordinate-level object category, the task is to retrieve the relevant document from a large, expertly-curated text corpus; to this end, we only allow non-expert supervision for learning to describe the image.
We assume that:
(1) the corpus dedicates a separate entry to each category, as is, for example, the case in encyclopedia entries for bird or plant species, etc.,
(2) there exist no paired data of images and documents or expert labels during training, and 
(3) to model a layperson's capabilities, we have access to general image and text understanding tools that do not use expert knowledge, such as image descriptions or language models. 

Given this definition, the task classifies as weakly-supervised in the taxonomy of learning problems.
We note that there are fundamental differences to related topics, such as image-to-text retrieval and unsupervised image classification. 
Despite a significant amount of prior work in image-to-text or text-to-image retrieval~\cite{peng2017overview,wang2017adversarial,zhen19deep,hu2019separated,he2019new}, the general assumption is that images and corresponding documents are paired for training a model. 
In contrast to unsupervised image classification, the difference is that here we are interested in \emph{semantically} labelling images using a secondary modality, instead of grouping similar images~\cite{asano2020self, caron2020unsupervised, vangansbeke2020scan}.

To the best of our knowledge, we are the first to tackle the task of fine-grained image recognition without expert supervision. 
Since the target corpus is not required during training, the search domain is easily extendable to any number of categories/species---an ideal use case when retrieving documents from dynamic knowledge bases, such as Wikipedia.
We provide extensive evaluation of our method and also compare to approaches in cross-modal retrieval, %
despite using significantly reduced supervision.

\definecolor{mycol}{rgb}{1, 0, 0}

\section{Related Work}\label{s:related}

In this paper, we address a novel problem (\ourtask). %
Next we describe in detail how it differs from related problems in the computer vision and natural language processing literature and summarise the differences with respect to how class information is used in \cref{tab:rw}.

\paragraph{Fine-Grained Recognition.}
\begin{wrapfigure}{r}{0.37\textwidth}
\makeatletter
\def\@captype{table}
\makeatother
\centering
\small
\begin{tabular}{lcc} %
	\toprule
	 & \multicolumn{2}{c}{\textbf{Class Information}} \\ 
	\cmidrule{2-3}
	\textbf{Task} & \textbf{Train} & \textbf{Test} \\
	\midrule
	FGVR & K & K \\
    ZSL & K & U \\
	GZSL  & K & K + U \\
	\ourtask & U & U \\
	\bottomrule
\end{tabular}
\vspace{0.5em}
\caption{Overview of related topics (K: known, U: unknown).}
\vspace{-1em}
\label{tab:rw}
\end{wrapfigure}

The goal of fine-grained visual recognition (FGVR) is categorising objects at sub-ordinate level, such as species of animals or plants~\cite{wah2011caltech,van2015building,van2018inaturalist,nilsback2008automated, kumar2012leafsnap}.
Large-scale annotated datasets %
require domain experts and are thus difficult to collect.
FGVR is more challenging than coarse-level image classification as it involves categories with fewer discriminative cues and fewer labeled samples.
To address this problem, supervised methods exploit side information such as part annotations~\cite{zhang2014part}, attributes~\cite{vedaldi2014understanding}, natural language descriptions~\cite{he2017fine}, noisy web data~\cite{krause2016unreasonable,xu2016webly,gebru2017fine} or humans in the loop~\cite{branson2010visual,deng2015leveraging,cui2016fine}.
Attempts to reduce supervision in FGVR are mostly targeted towards eliminating auxiliary labels, \eg part annotations~\cite{zheng2017learning,simon2015neural,ge2019weakly,huang2020interpretable}.
In contrast, our goal is fine-grained recognition without access to \emph{categorical} labels during training.
Our approach only relies on side information (captions) provided by laymen and is thus unsupervised from the perspective of ``expert knowledge''.

\paragraph{Zero/Few Shot Learning.}

Zero-shot learning (ZSL) is the task of learning a classifier for unseen classes~\cite{xian2018zero}.
A classifier is generated from a description of an object in a secondary modality, mapping semantic representations to class space in order to recognize said object in images~\cite{socher2013zero}.
Various modalities have been used as auxiliary information: word embeddings~\cite{frome13devise:,xian2016latent}, hierarchical embeddings~\cite{kampffmeyer2019rethinking}, attributes~\cite{5206772,akata2015label} or Wikipedia articles~\cite{elhoseiny2017link,zhu2018generative,elhoseiny2016write,qiao2016less}.
Most recent work uses generative models conditioned on class descriptions to synthesize training examples for unseen categories~\cite{long2017zero,kodirov2017semantic,felix2018multi,xian2019f,Vyas2020LeveragingSA,xian2018feature}.
The multi-modal and often fine-grained nature of the standard and generalised (G)ZSL task renders it related to our problem.
However, different from the (G)ZSL settings our method uses neither class supervision during training nor image-document pairs as in~\cite{elhoseiny2017link,zhu2018generative,elhoseiny2016write,qiao2016less}.

\paragraph{Cross-Modal and Information Retrieval.}
While information retrieval deals with extracting information 
from document collections~\cite{manning2008introduction}, cross-modal retrieval aims at retrieving relevant information across various modalities, \eg image-to-text or vice versa.
One of the core problems in information retrieval is ranking documents given some query, with a classical example being Okapi BM25~\cite{robertson1995okapi}.
With the advent of transformers~\cite{vaswani2017attention} and BERT~\cite{Devlin2019BERTPO}, state-of-the-art document retrieval is achieved in two-steps; an initial ranking based on keywords followed by computationally intensive BERT-based re-ranking~\cite{nogueira2019passage,nogueira2020document,yilmaz2019cross,macavaney2019cedr}.
In cross-modal retrieval, the common approach is to learn a shared representation space for multiple modalities~\cite{peng2017overview, andrew2013deep,wang2015large, peng2016cross,peng2017ccl,wang2017adversarial,zhen19deep,hu2019separated,he2019new}.
In addition to paired data in various domains, some methods also exploit auxiliary semantic labels; for example, the Wikipedia benchmark~\cite{pereira2013role} provides broad category labels such as \emph{history}, \emph{music}, \emph{sport}, etc.

We depart substantially from the typical assumptions made in this area.
Notably, with the exception of \cite{he2019new,Wang09}, this setting has not been explored in fine-grained domains, but generally targets higher-level content association between images and documents.
Furthermore, one major difference between our approach and cross-modal retrieval, including~\cite{he2019new,Wang09}, is that we do not assume paired data between the input domain (images) and the target domain (documents).
We address the lack of such pairs using an intermediary modality (captions) that allows us to perform retrieval directly in the text domain.

\paragraph{Natural Language Inference (NLI) and Semantic Textual Similarity (STS).}
Also related to our work, in natural language processing, the goal of the NLI task is to recognize textual entailment, \ie given a pair of sentences (premise and hypothesis), the goal is to label the hypothesis as entailment (true), contradiction (false) or neutral (undetermined) with respect to the premise~\cite{snli_emnlp2015, williams2018broad}.
STS measures the degree of semantic similarity between two sentences~\cite{agirre-etal-2012-semeval, agirre-etal-2013-sem}. 
Both tasks play an important role in semantic search and information retrieval and are currently dominated by the transformer architecture~\cite{vaswani2017attention, Devlin2019BERTPO, liu2019roberta, reimers2019sentence}.
Inspired by these tasks, we propose a sentence similarity regime that is domain-specific, paying attention to fine-grained semantics.

\section{Method}\label{s:method}

We introduce the problem of layperson-to-expert visual entity recognition (\ourtask), which we address via image-based document retrieval. 
Formally, we are given a set of images $x_i \in \mathcal{I}$ to be labelled given a corpus of expert documents $D_j \in \mathcal{D}$, where each document corresponds to a fine-grained image category and there exist $K = |D|$ categories in total. 
As a concrete example, $\mathcal{I}$ can be a set of images of various bird species and $\mathcal{D}$ a bird identification corpus constructed from specialized websites (with one article per species). 
Crucially, the pairing of $x_i$ and $D_j$ is not known, \ie no expert task supervision is available during training.
Therefore, the mapping from images to documents cannot be learned directly but can be discovered through the use of non-expert image descriptions $\mathcal{C}_i$ for image $x_i$. 

\begin{figure*}[!t]
\centering
\includegraphics[width=0.95\textwidth,trim=0 10.8cm 2cm 0]{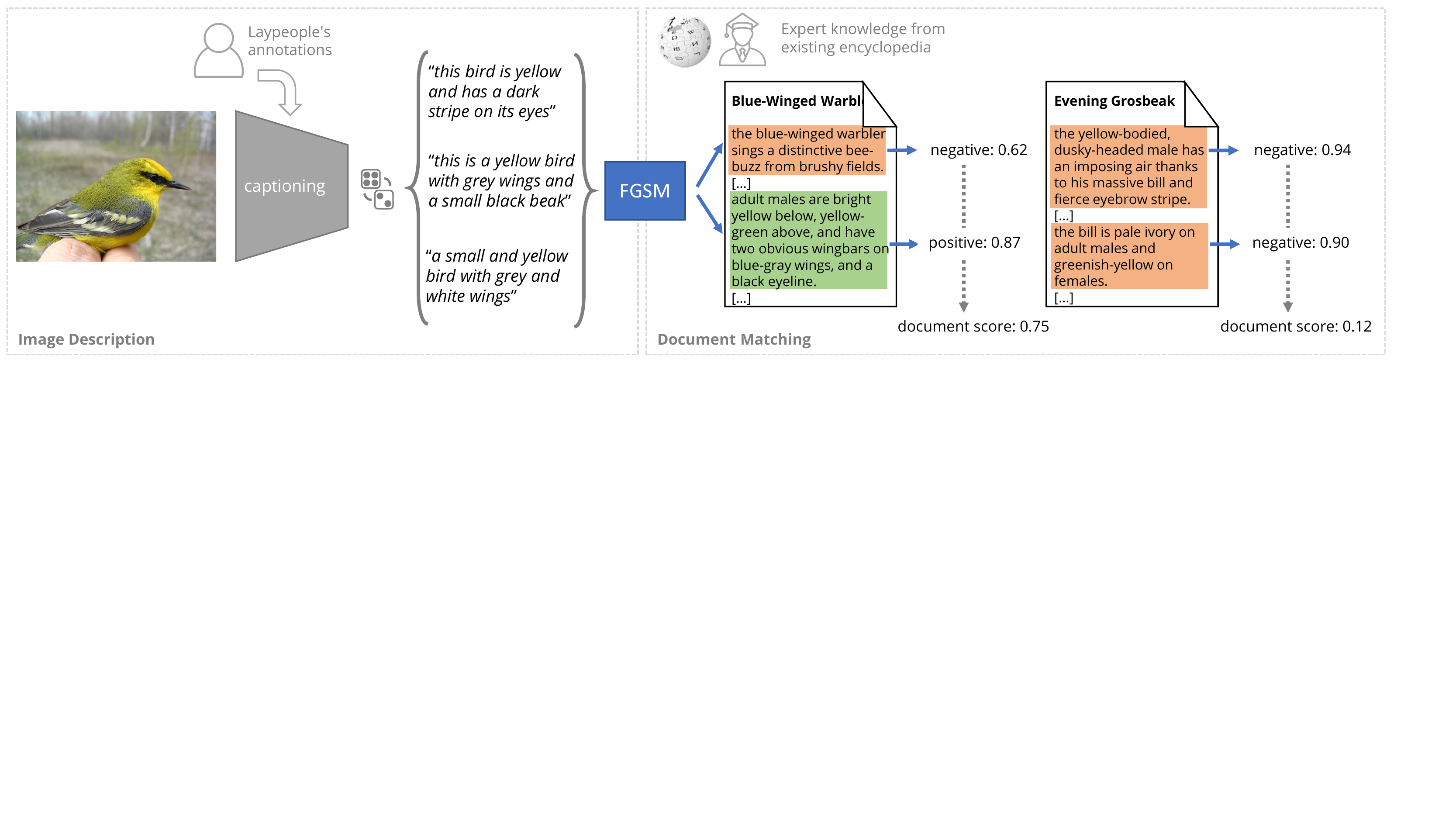}
\caption{\textbf{Overview.} We train a model for fine-grained sentence matching (FGSM) using layerperson's annotations, \ie class-agnostic image descriptions. At test time, we score documents from a relevant corpus and use the top-ranked document to label the image.}
\label{fig:method}
\end{figure*}

Our method consists of three distinct parts.
First, we learn, using ``layperson's supervision'', an image captioning model that uses simple color, shape and part descriptions.
Second, we train a model for Fine-Grained Sentence Matching (FGSM).
The FGSM model takes as input a pair of sentences and predicts whether they are descriptions of the same object. 
Finally, we use the FGSM to score the documents in the expert corpus via voting.
As there is one document per class, the species corresponding to the highest-scoring document is returned as the final class prediction for the image.
The overall inference process is illustrated in~\cref{fig:method}.

\subsection{Fine-grained Sentence Matching}
\label{subsec:fgsm}
The overall goal of our method is to match images to expert documents\,---\,however, in absence of paired training data, learning a cross-domain mapping is not possible. 
On the other hand, describing an image is an easy task for most humans, as it usually does not require domain knowledge. 
It is therefore possible to leverage image descriptions as an intermediary for learning to map images to an expert corpus. 

To that end, the core component of our approach is the FGSM model $f(c_1, c_2) \in \mathbb{R}$ that scores the visual similarity of two descriptions $c_1$ and $c_2$. 
We propose to train $f$ in a manner similar to the textual entailment (NLI) task in natural language processing.
The difference to NLI is that the information that needs to be extracted here is fine-grained and domain-specific \eg~``\textit{a bird with blue wings}'' vs.~``\textit{this is a uniformly yellow bird}''.
Since we do not have annotated sentence pairs for this task, we have to create them synthetically. 
Instead of the terms \texttt{entailment} and \texttt{contradiction}, here we use \texttt{positive} and \texttt{negative} to emphasize that the goal is to find matches (or mismatches) between image descriptions.

We propose to model $f$ as a sentence encoder, performing the semantic comparison of $c_1, c_2$ in embedding space. 
Despite their widespread success in downstream tasks, most transformer-based language models are notoriously bad at producing semantically meaningful sentence embeddings~\cite{reimers2019sentence,li2020sentence}.
We thus follow \cite{reimers2019sentence} in learning an appropriate textual similarity model with a Siamese architecture built on a pre-trained language transformer. 
This also allows us to leverage the power of large language models while maintaining efficiency by computing an embedding for each input independently and only compare embeddings as a last step.
To this end, we compute a similarity score for $c_1$ and $c_2$ as $f(c_1,\,c_2) = h\left(\left[\phi_1;\,\phi_2;\, |\phi_1 - \phi_2|\right]\right)$,
where $[\cdot]$ denotes concatenation, and $h$ and $\phi$ are lightweight MLPs operating on the average-pooled output of a large language model $\bert(\cdot)$ with the shorthand notation $\phi_1 = \phi(\bert(c_1))$.
\paragraph{Training.}
One requirement is that the FGSM model should be able to identify \emph{fine-grained} similarities between pairs of sentences.
This is in contrast to the standard STS and NLI tasks in natural language understanding which determine the relationship (or degree of similarity) of a sentence pair on a \emph{coarser} semantic level.
Since our end-goal is visual recognition, we instead train the model to emphasize visual cues and nuanced appearance differences.

Let $\mathcal{C}_i$ be the set of human-annotated descriptions for a given image $x_i$.
Positive training pairs are generated by exploiting the fact that, commonly, each image has been described by multiple annotators; for example in CUB-200~\cite{wah2011caltech} there are $|\mathcal{C}_i| = 10$ captions per image.
Thus, each pair (from $\mathcal{C}_i \times \mathcal{C}_i$) of descriptions of the same image can be used as a positive pair.
The negative counterparts are then sampled from the complement $\bar{\mathcal{C}}_i = \bigcup_{l \neq i}\mathcal{C}_l$, \ie among the available descriptions for all other images in the dataset.
We construct this dataset with an equal amount of samples for both classes and train $f$ with a binary cross entropy loss.

\paragraph{Inference.}
During inference the sentence embeddings $\phi$ for each sentence in each document can be precomputed and only $h$ needs to be evaluated dynamically given an image and its corresponding captions, as described in the next section.
This greatly reduces the memory and time requirements.

\subsection{Document Scoring}
Although trained from image descriptions alone, the FGSM model can take any sentence as input and, at test time, we use the trained model to score sentences from an expert corpus against image descriptions.
Specifically, we assign a score $z_{ij} \in \mathbb{R}$ to each expert document $D_j$ given a set of descriptions for the $i$-th image: 
\begin{equation}\label{eq:ranking}
    z_{ij} = \frac{1}{|\mathcal{C}_i\times D_j|} \sum_{(c, s) \in \mathcal{C}_i\times D_j}{f(c, s)} ,
\end{equation}
Since there are several descriptions in $\mathcal{C}_i$ and sentences in $D_j$, we compute the final document score as an average of individual predictions (scores) of all pairs of descriptions and sentences.
Aggregating scores across the whole corpus $\mathcal{D}$, 
we can then compute the probability 
$p(D_j \:\vert\: x_i) = \frac{e^{-z_{ij}}}{\sum_{k} e^{-z_{ik}}}$ 
of a document $D_j \in \mathcal{D}$ given image $x_i$
and assign the document (and consequently class) with the highest probability to the image.

\subsection{Bridging the Domain Gap}
\label{subsec:domaingap}
While training the FGSM model, we have so far only used laypersons' descriptions, disregarding the expert corpus. 
However, we can expect the documents to contain significantly more information than visual descriptions. 
In the case of bird species, encyclopedia entries usually also describe behavior, migration, conservation status, etc. 
In this section, we thus employ two mechanisms to bridge the gap between the image descriptions and the documents. 
\paragraph{Neutral Sentences.}
We introduce a third, \texttt{neutral} class to the classification problem, designed to capture sentences that do not provide relevant (visual) information. We generate neutral training examples by pairing an image description with sentences from the documents (or other descriptions) that do not have any \emph{nouns} in common. Instead of binary cross entropy, we train the three-class model (positive/neutral/negative) with softmax cross entropy.

\paragraph{Score Distribution Prior.}
Despite the absence of paired training data, we can still impose priors on the document scoring. 
To this end, we consider the probability distribution $p(\mathcal{D} \:\vert\: x)$ over the entire corpus $\mathcal{D}$ given an image $x$ in a training batch $\mathcal{B}$. 
We can then derive a regularizer $R(\mathcal{B})$ that operates at batch-level:
\begin{equation}
    R(\mathcal{B}) = \sum_{x \in \mathcal{B}} \Big( -\langle p(\mathcal{D} \:\vert\: x),\ p(\mathcal{D} \:\vert\: x) \rangle \; \\ +  \sum_{x' \in \mathcal{B} \setminus x } \langle p(\mathcal{D} \:\vert\: x),\ p(\mathcal{D} \:\vert\: x') \rangle \Big)
\end{equation}
where $\langle \cdot, \cdot \rangle$ denotes the inner product of two vectors. The intuition of the two terms of the regularizer is as follows. 
$\langle p(D\:\vert\:x),\: p(D\:\vert\:x) \rangle$ is maximal when the distribution assigns all mass to a single document. 
Since the score $z_{ij}$ is averaged over all captions of one image, this additionally has the side effect of encouraging all captions of one image to vote for the same document. 
The second term of $R(\mathcal{B})$ then encourages the distributions of two different images to be orthogonal, favoring the assignment of images uniformly across all documents. 

Since $R(\mathcal{B})$ requires evaluation over the whole document corpus for every image, we first pre-train $f$, including the large transformer model $T$, (c.f.~\cref{subsec:fgsm}). 
After convergence, we extract sentence features for all documents and image descriptions and train only the MLPs $\phi$ and $h$ 
with $\mathcal{L} + \lambda R$, where $\lambda$ balances the 3-class cross entropy loss $\mathcal{L}$ and the regularizer. 

\section{Experiments}

We validate our method empirically for bird and plant identification.
To the best of our knowledge, we are the first to consider this task, thus in absence of state-of-the-art methods, we ablate the different components of our model and compare to several strong baselines.

\subsection{Datasets and Experimental Setup}

\paragraph{Datasets.}
We evaluate our method on Caltech-UCSD Birds-200-2011 (CUB-200)~\cite{wah2011caltech} and the Oxford-102 Flowers (FLO) dataset~\cite{Nilsback06}.
For both datasets, \citet{reed2016learning} have collected several visual descriptions per image by crowd-sourcing to non-experts on Amazon Mechanical Turk (AMT).
We further collect for each class a corresponding expert document from specialised websites, such as AllAboutBirds\footnote{\url{https://allaboutbirds.com}} (AAB) and Wikipedia.
\begin{table*}[!t]
\begin{center}
	\small  
	\tabcolsep=0.15cm
    \renewcommand*{\arraystretch}{0.95}
	\begin{tabular}{l@{\hskip 0.5cm}rrr@{\hskip 0.5cm}rrr}
		\toprule
		& \multicolumn{3}{c@{\hskip 0.5cm}}{\textbf{CUB-200}} &  \multicolumn{3}{c}{\textbf{FLO}} \\
		\cmidrule(l{-0.1cm}r{0.2cm}){2-4}
		\cmidrule(r{-0.2cm}r){5-7}
		\textbf{Method}  & \textbf{top-1}$\uparrow$ & \textbf{top-5}$\uparrow$ & \textbf{MR}$\downarrow$ & \textbf{top-1}$\uparrow$ & \textbf{top-5}$\uparrow$ & \textbf{MR}$\downarrow$ \\
		\midrule
		random guess       & 0.5 &  2.5 &100.0 & 0.9 &  4.9 & 51.0  \\
		SRoBERTa-STSb~\cite{reimers2019sentence} (no-ft) & 1.3 & 6.4 & 73.4  &  1.1 & 7.7 & 45.2 \\  %
		SRoBERTa-NLI~\cite{liu2019roberta} (no-ft)  & 1.9 & 5.3 & 81.3 & 0.9 & 5.7 & 48.2  \\ %
		Okapi BM25~\cite{robertson1995okapi} & 1.0 & 7.5 & 78.2  & 1.6 & 8.0 & 43.9 \\ %
		TF-IDF~\cite{jones1972statistical} & 2.2 & 9.7 & 72.1  & 1.4 & 5.0 & 45.2 \\ %
		RoBERTa~\cite{liu2019roberta}  & 4.3 & 16.6 & 44.6  &  1.1 & 9.6 & 42.6 \\
		\midrule
		ours   & \textbf{7.9} & \textbf{28.6} & \textbf{31.9}  &  \textbf{6.2} & \textbf{14.2} & \textbf{39.7}   \\   %
		\bottomrule
	\end{tabular}
\end{center}
\caption{\textbf{Comparison to baselines.} We report the retrieval performance of our method on CUB-200 and Oxford-102 Flowers (FLO) and compare to various strong baselines. }
\label{tab:retrieval}
\end{table*}

\paragraph{Setup.}
We use the image-caption pairs to train two image captioning models: ``Show, Attend and Tell'' (SAT)~\cite{xu2015show}
and AoANet~\cite{huang2019attention}.
Unless otherwise specified, we report the performance of our model based on their ensemble, \ie combining predictions from both models.
As the backbone $T$ of our sentence transformer model, we use RoBERTa-large~\cite{liu2019roberta} fine-tuned on NLI and STS datasets using the setup of~\cite{reimers2019sentence}.
Please see the appendix for further implementation, architecture, dataset and training details.

We use three metrics to evaluate the performance on the benchmark datasets.
We compute top-1 and top-5 per-class retrieval accuracy and report the overall average.
Additionally, we compute the mean rank (MR) of the target document for each class.
Here, retrieval accuracy is identical to classification accuracy, since there is only a single relevant article per category.

\subsection{Baseline Comparisons}\label{s:e:baselines}

Since this work is the first to explore the mapping of images to expert documents without expert supervision, we compare our method to several strong baselines (\cref{tab:retrieval}).

Our FGSM performs text-based retrieval, we evaluate current text retrieval systems. 
\textbf{TF-IDF:}
Term frequency-inverse document frequency (TF-IDF) is 
widely used for unsupervised document retrieval \cite{jones1972statistical}.
For each image, we use the predicted captions as queries and use the TF-IDF textual representation for document ranking instead of our model.
We empirically found the cosine distance and $n$-grams with $n={2,3}$ to perform best for TF-IDF.
\textbf{BM25:}
Similar to TF-IDF, BM25~\cite{robertson1995okapi} is another common measure for document ranking based on $n$-gram frequencies.
We use the BM25 Okapi implementation from the python package \texttt{rank-bm25} with default settings.
\textbf{RoBERTa:}
One advantage of processing caption-sentence pairs with a Siamese architecture, such as SBERT/SRoBERTa~\cite{reimers2019sentence}, is the reduced complexity.  
Nonetheless, we have trained a transformer baseline for text classification, using the same backbone~\cite{liu2019roberta}, concatenating each sentence pair with a \texttt{SEP} token and training as a binary classification problem.
We apply this model to score documents, instead of FGSM, aggregating scores at sentence-level.
\textbf{SRoBERTa-NLI/STSb:} Finally, to evaluate the importance of learning \emph{fine-grained} sentence similarities, we also measure the performance of the same model trained only on the NLI and STSb benchmarks~\cite{reimers2019sentence}, without further fine-tuning.
Following \cite{reimers2019sentence}
we rank documents based on the cosine similarity between the caption and sentence embeddings.

Our method outperforms all bag-of-words and learned baselines.
Approaches such as TF-IDF and BM25 are very efficient, albeit less performant than learned models.
Notably, the closest in performance to our model is the transformer baseline (RoBERTa), which comes at a large computational cost ($347$ sec vs.~$0.55$ sec for our model per image on CUB-200).

\begin{table}
\parbox{.47\linewidth}{
\begin{center}
    \small
	\tabcolsep=0.12cm
    \renewcommand*{\arraystretch}{0.95}
	\begin{tabular}{lrrr}
		\toprule
		\textbf{Method} & \textbf{top-1}$\uparrow$ & \textbf{top-5}$\uparrow$ & \textbf{MR}$\downarrow$ \\
		\midrule
		user interaction         & 11.9 & 37.5 & 24.8 \\ %
		\midrule\midrule
		FGSM + cosine       & 4.5 & 17.8 & 35.5 \\ %
		\midrule
		FGSM w/ SAT          & 4.3 & 15.0 & 42.9 \\ %
		FGSM w/ AoANet       & 5.7 & 20.8 & 38.3 \\ %
		FGSM w/ ensemble     & 5.9 & 20.0 & 36.1 \\ %
		\midrule
		FGSM $+\; R(\mathcal{B})$ [2-cls]   &  7.4 & 24.6 & \textbf{29.9}  \\ %
		FGSM $+\; R(\mathcal{B})$ [3-cls]    & \textbf{7.9} & \textbf{28.6} & 31.9 \\ %
		
		\bottomrule
	\end{tabular}
\end{center}
\caption{\textbf{Ablation and user study.} On CUB-200 we evaluate scoring functions, captioning models and the regularizer $R(\mathcal{B})$.}
\label{tab:ablation}
}
\hfill
\parbox{.47\linewidth}{
\begin{center}
    \small
	\tabcolsep=0.12cm
    \begin{tabular}{l crrr}
        \toprule
         \textbf{Method} & \textbf{sup.} & \textbf{top-1}$\uparrow$ & \textbf{top-5}$\uparrow$ & \textbf{MR}$\downarrow$ \\
         \midrule
         random guess & \xmark & 2.0 & 10.0 & 25.0 \\
         ViLBERT~\cite{lu2019vilbert} & \xmark & 3.5 & 14.8 & 20.2 \\ %
         TF-IDF~\cite{jones1972statistical}& \xmark & 7.2 & 28.6 & 18.9 \\ %
         CLIP~\cite{radford2021learning} & \cmark & 10.0 & 32.9 & 14.0 \\ %
         DSCMR~\cite{zhen19deep} & \cmark & 13.5 & 34.7 & 15.2 \\
         \midrule
         ours & \xmark & \textbf{20.9} & \textbf{50.7} & \textbf{9.6} \\
         \bottomrule
    \end{tabular}
\end{center}

\caption{\textbf{Comparison to cross-media retrieval.} We evaluate the performance of methods on the ZSL split of CUB-200. Our method performs favorably against existing approaches trained with more supervision.}
\label{tab:retrieval_sota}
}
\end{table}

\subsection{Ablation \& User Interaction}
We ablate the different components of our approach in \cref{tab:ablation}.
We first investigate the use of a different scoring mechanism, \ie the cosine similarity between the embeddings of $c$ and $s$ as in \cite{reimers2019sentence}; we found this to perform worse (FGSM + cosine).
We also study the influence of the captioning model on the final performance. %
We evaluate captions obtained by two methods, SAT~\cite{xu2015show} and AoANet~\cite{huang2019attention}, as well as their ensemble.
The ensemble  
improves performance thanks to higher variability in the image descriptions.
Next, we evaluate the performance of our model after the final training phase, with the proposed regularizer and the inclusion of neutral pairs (\cref{subsec:domaingap}). 
$R(\mathcal{B})$ imposes prior knowledge about the expected class distribution over the dataset and thus stabilizes the training, resulting in improved performance ([2-cls]). 
Further, through the regularizer and neutral sentences ([3-cls]), FGSM is exposed to the target corpus during training, which helps reduce the domain shift during inference 
compared to training on image descriptions alone (FGSM w/ ensemble).

Finally, our method enables user interaction, \ie allowing a user to directly enter own descriptions, replacing the automatic description model.
In~\cref{tab:ablation} we have simulated this by evaluating with ground-truth instead of predicted descriptions.
Naturally, we find that human descriptions indeed perform better, though the performance gap is small.
We attribute this gap to a much higher diversity in the human annotations.
Current image captioning models still have diversity issues, %
which also explains why our ensemble variant improves the results.

\begin{figure*}[!t]
\centering
\includegraphics[width=0.94\textwidth,trim=0cm 5cm 1cm 0]{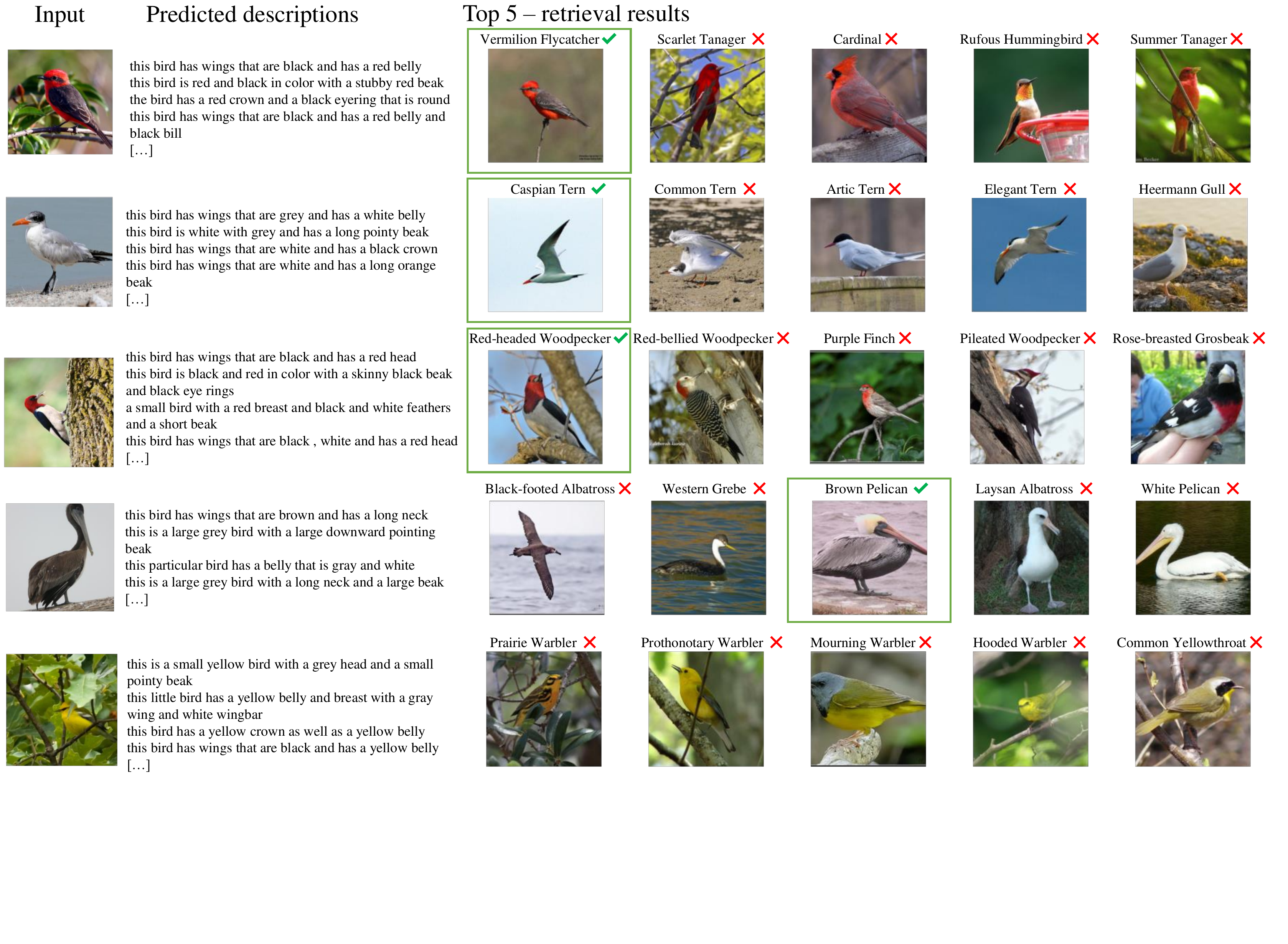}
\caption{\textbf{Qualitative Results (CUB-200).} We show examples of input images and their predicted captions, followed by the top-5 retrieved documents (classes). For illustration purposes, we show a random image for each document; the image is not used for matching. %
}\label{fig:qualitative}
\vspace{-.5em}
\end{figure*}

\subsection{Comparison with Cross-Modal Retrieval}

Since the nature of the problem presented here is in fact cross-modal, we adapt a representative method, DSCMR~\cite{zhen19deep}, to our data to compare to the state of the art in cross-media retrieval.
We note that such an approach requires image-document pairs as training samples, thus using more supervision than our method.
Instead of using image descriptions as an intermediary for retrieval, DSCMR thus performs retrieval monolithically, mapping the modalities in a shared representation space. 
We argue that, although this is the go-to approach in broader category domains, it may be sub-optimal in the context of fine-grained categorization. 

Since in our setting each category (species) is represented by a single article, in the scenario that a supervised model sees \emph{all} available categories during training, the cross-modal retrieval problem degenerates to a classification task.
Hence, for a meaningful comparison, we train both our model and DSCMR on the CUB-200 splits for ZSL~\cite{xian2018zero} to evaluate on 50 \emph{unseen} categories.
We report the results in \cref{tab:retrieval_sota}, including a TF-IDF baseline on the same split. 
Despite using no image-documents pairs for training, our method still performs significantly better.

Additionally, we compare to representative methods from the vision-and-language representation learning space.
ViLBERT~\cite{lu2019vilbert} is a multi-modal transformer model capable of learning joint representations of visual content and natural language.
It is pre-trained on 3.3M image-caption pairs with two proxy tasks.
We use their multi-modal alignment prediction mechanism to compute the alignment of the sentences in a document to a target image, similar to ViLBERT's zero-shot experiments. 
The sentence scores are averaged to get the document alignment score and the document with the maximum score is chosen as the class.
Finally, we compare to CLIP~\cite{radford2021learning}, that learns a multimodal embedding space from 400M image-text pairs. 
CLIP predicts image and sentence embeddings with separate encoders. 
For a target image we score each sentence using cosine similarity and average across the document for the final score. 
CLIP's training data is not public, but we find that there is a high possibility it does indeed contain expert labels as removing class names from documents hurts its performance.

\subsection{Qualitative Results}
In \cref{fig:qualitative}, we show qualitative retrieval results.
The input image is shown on the left followed by the predicted descriptions.
We then show the top-5 retrieved documents/classes together with an example image for the reader.
Note that the example images are not used for matching, as the FGSM module operates on text only.
We find that in most cases, even when the retrieved document does not match the ground truth class, the visual appearance is still similar.
This is especially noticeable in families of birds for which discriminating among individual species is considered to be particularly difficult even for humans, \eg warblers (last row).

\section{Discussion}
Like with any method that aims to reduce supervision, our method is not perfect. 
There are multiple avenues where our approach can be further optimized.

First, we observe that models trained for image captioning tend to produce short sentences that lack distinctiveness, focusing on the major features of the object rather than providing detailed fine-grained descriptions of the object's unique aspects.  
We believe there is a scope for improvement if the captioning models could extensively describe each different part and attribute of the object.
We have tried to mitigate this issue by using an ensemble of two popular captioning networks.
However, using multiple models and sampling multiple descriptions may lead to redundancy.
Devising image captioning models that produce diverse and distinct fine-grained image descriptions may provide improved performance on CLEVER task;
there is an active area of research~\cite{wang2020compare, wang2020OnDi} that is looking into this problem.

Second, the proposed approach to scoring a document given an image uses \emph{all} the sentences in the document classifying them as positive, negative or neutral with respect to each input caption.
Given that the information provided by an expert document might be noisy, \ie not necessarily related to the \emph{visual} domain, 
it is likely worthwhile to develop a filtering mechanism for relevancy, effectively using only a subset of the sentences for scoring.

Finally, in-domain regularization results in a significant performance boost (\cref{tab:ablation}), which implies that the CLEVER task is susceptible to the domain gap between laypeople's descriptions and the expert corpus.
Language models such as BERT/RoBERTa partially address this problem already by learning general vocabulary, semantics and grammar during pre-training on large text corpora, enabling generalization to a new corpus without explicit training.
However, further research in reducing this domain gap seems worthwhile.

\section{Conclusion}
We have shown that it is possible to address fine-grained image recognition without the use of expert training labels by leveraging existing knowledge bases, such as Wikipedia.
This is the first work to tackle this challenging problem, with performance gains over the state of the art on cross-media retrieval, despite their training with image-document pairs.
While humans can easily access and retrieve information from such knowledge bases, \ourtask remains a challenging learning problem that merits future research.  

\section*{Acknowledgments}
S. C. is supported by a scholarship sponsored by Facebook.
I. L. is supported by the European Research Council (ERC) grant IDIU-638009 and EPSRC VisualAI EP/T028572/1.
C. R. is supported by Innovate UK (project 71653) on behalf of UK Research and Innovation (UKRI) and ERC grant IDIU-638009.
A. V. is supported by ERC grant IDIU-638009.
We thank Andrew Brown for valuable discussions. 
\bibliography{{vedaldi_general,references}}
\end{document}